\documentclass[11pt]{article}

\usepackage[preprint]{acl}

\usepackage{times}
\usepackage{latexsym}

\usepackage[T1]{fontenc}

\usepackage[utf8]{inputenc}

\usepackage{microtype}

\usepackage{inconsolata}

\usepackage{graphicx}
\usepackage{adjustbox}
\usepackage[english]{babel}
\usepackage{xcolor}
\usepackage{listings}

\title{Neuro-symbolic Training for Spatial Reasoning over Natural Language}

\author{Tanawan Premsri \\
  Department of Computer Science\\
Michigan State University \\
  \texttt{premsrit@msu.edu} \\\And
  Parisa Kordjamshidi \\
  Department of Computer Science\\
Michigan State University \\
  \texttt{kordjams@msu.edu} \\}

\begin{document}

\maketitle

\begin{abstract}

Spatial reasoning based on natural language expressions is essential for everyday human tasks. 
This reasoning ability is also crucial for machines to interact with their environment in a human-like manner. 
However, recent research shows that even state-of-the-art language models struggle with spatial reasoning over text, especially when facing nesting spatial expressions. 
This is attributed to not achieving the right level of abstraction required for generalizability.
To alleviate this issue, we propose training language models with neuro-symbolic techniques that exploit the spatial logical rules as constraints, providing additional supervision to improve spatial reasoning and question answering.
Training language models to adhere to spatial reasoning rules guides them in making more effective and general abstractions for transferring spatial knowledge to various domains. 
We evaluate our approach on existing spatial question-answering benchmarks. 
Our results indicate the effectiveness of our proposed technique in improving language models in complex multi-hop spatial reasoning over text.

\end{abstract}

\section{Introduction}

\begin{figure}[t]
\begin{center}
\includegraphics[width=\linewidth]{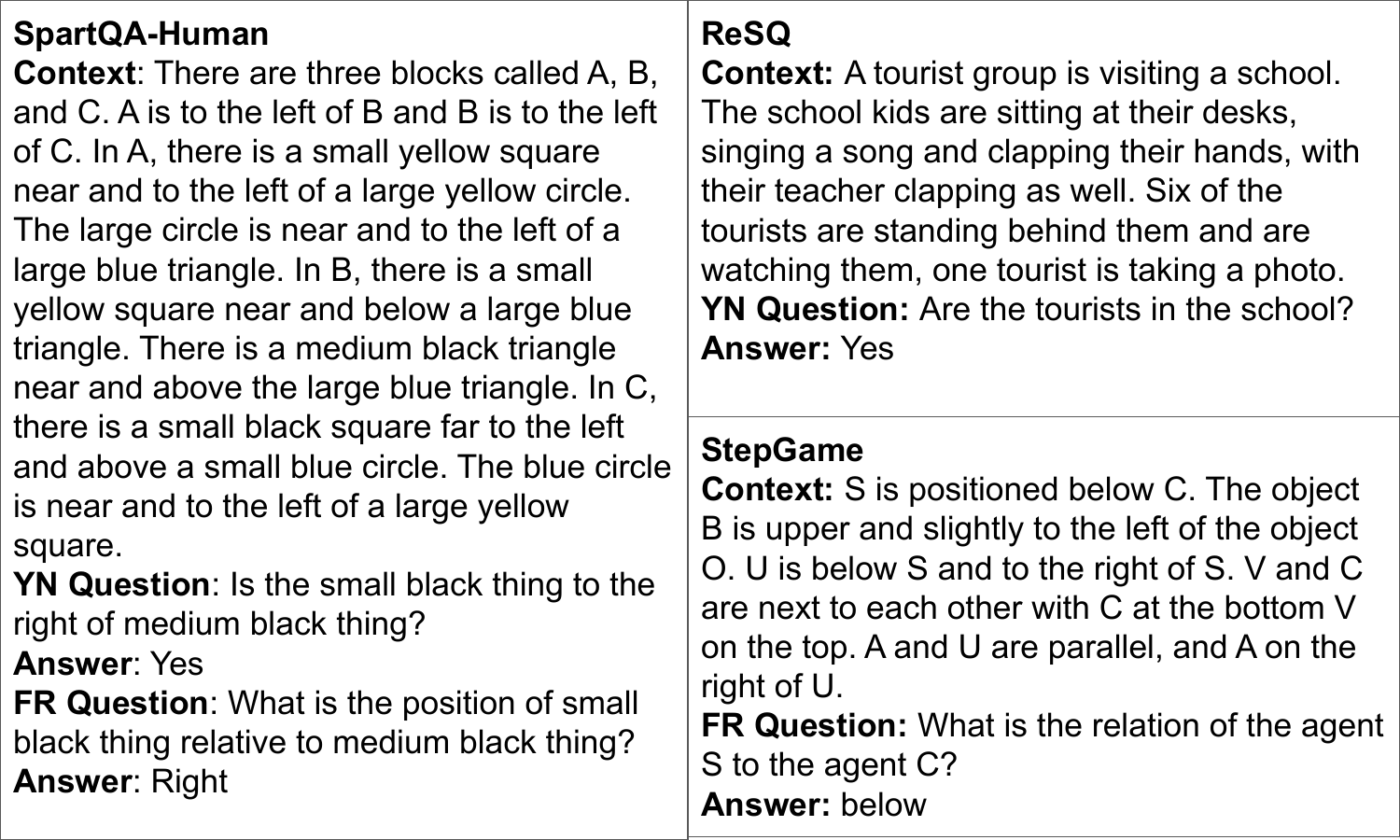} 
\caption{Examples of story and questions in Spatial Question Answering (SQA). YN is Yes-No, and FR is Find relation type of questions.}
\label{fig.1}
\end{center}
\end{figure}

Spatial language conveys information about the location and translocation of objects/entities in space. Reasoning over spatial language is required for natural language processing systems to answer "Where" questions and identify the relative spatial relations between objects explained in the text, such as left and right. Figure~\ref{fig.1}, demonstrates a spatial language comprehension and question-answering problem. 
Understanding spatial language turns out to be crucial for many AI applications, including question-answering systems, grounding language into visual modality ~\cite{zhang2021navigation,zhang2022lovis, liu2023visual}, navigating robots~\cite{sisbot2007spatial,10.1145/3623809.3623830} and even in more specific fields such as medical domain~\cite{atif2007generic, DATTA2020103473, gong20233dsamadapter}.

Despite the high performance of Large Language models (LLMs) in many NLP and vision problems~\cite{brown2020language, openai2023GPT4}, recent studies highlight the shortcomings in complex spatial reasoning over text~\cite{bang2023multitaskmultilingualmultimodalevaluation, yang2023coupling, mirzaee2023disentangling,liu2023visual, chen2024spatialvlm}. 
This paper focuses on complex, that is, multi-hop spatial reasoning over text. However, our research findings can advance complex reasoning over text in general. Our main idea is to exploit spatial logical rules of reasoning in training language models. 

Explicit logic, in combination with LLMs, has been used in some related works. In particular, \citealt{yang2023coupling} integrated explicit symbolic reasoning with LLMs in a pipeline. 
The LLMs map natural language to symbolic representation, and this formal representation is passed to an Answer Set Programming (ASP) solver to conduct logical reasoning. 
Similarly, \citealt{mirzaee2023disentangling} passed formal representations obtained from LLMs to Prolog to conduct logical reasoning and answer spatial questions.
These approaches rely on external reasoning tools and require accurate formal representations generated at inference time. Obtaining such formal representations for realistic domain natural language is hard and, as shown in~\cite{mirzaee2023disentangling}, limits their effectiveness only to synthetic benchmarks.

Here, we aim to address the limitations of utilizing spatial logic in realistic domains. 
In our proposed technique, we fine-tune language models using spatial logical rules to guide them in capturing the level of abstraction needed for general spatial reasoning.
In particular, we train the models to minimize both cross-entropy loss and violation from logical constraints.
We hypothesize that incorporating logical knowledge as a source of supervision, even using synthetic data, enriches the models' abstract reasoning over spatial concepts, further enhancing their generalizability. 
A key advantage of our approach is that it does not require full coverage of logical knowledge. 
Any partially available logical knowledge can be utilized during training.
Additionally, formal knowledge representation is not needed at inference time, which is crucial because generating formal representations during inference is a challenging task in itself. 
Our technique also avoids reliance on external reasoning tools, which can add computational complexity in real-time applications.

We use the available three benchmarks for spatial question answering, SpartQA-Human~\cite{mirzaee2021SpartQA}, ReSQ~\cite{mirzaee2022transfer}, and StepGame~\cite{shi2022StepGame}
to evaluate our proposed method. 
Our improved results confirm our hypothesis about the impact of neuro-symbolic training on generalizability.  
The contribution of this paper can be summarized as follows\footnote{The code, trained models, and data are publicly available at \url{https://github.com/HLR/SpaRTUNQChain}}, 
{\textbf{(1)}} We exploit the available logical spatial knowledge in tuning language models to improve their spatial reasoning.
{\textbf{(2)}} We provide extensive experimental results and analysis on the existing benchmarks. Our results consistently confirm the effectiveness of our proposed approach in both encoder-based and generative language models and their generalizability for transferring knowledge to different domains;
{\textbf{(3)}} We show that tuning smaller models with our proposed method generalizes better on out-of-domain and complex reasoning compared to using prompt-engineered, very large language models.

\section{Related Work}
Investigating the reasoning capability of NLP models has become an important research topic given the instability in the performance of LLMs~\cite{gendron2023large, valmeekam2023large, feng2024naturally, chen2024spatialvlm, bang2023multitaskmultilingualmultimodalevaluation}. 
Typically, this ability is assessed through question-answering (QA) benchmarks~\cite{weston2015aicomplete, ZHOU2020275, tan2023benchmarking}. 
However, fewer studies are focused on spatial reasoning over text.
Recent benchmarks, such as SpaRTUN~\cite{mirzaee2022transfer} and StepGame~\cite{shi2022StepGame, li2024advancingspatialreasoninglarge}, have been developed to address this issue by providing comprehensive evaluation resources for spatial question answering.  
The current results using the existing benchmarks indicate that LLMs struggle with spatial reasoning over text, even on synthetic datasets~\cite{mirzaee2022transfer, yang2023coupling, bang2023multitaskmultilingualmultimodalevaluation}.

Several works have attempted to improve LLMs’ spatial reasoning capabilities.
~\citealt{mirzaee2022transfer} fine-tunes LLMs with synthetic data, which improves their multi-hop reasoning abilities, even when applied to realistic domains. 
Another approach in~\cite{shi2022StepGame} designed a specialized memory network based on a recurrent neural network for deep reasoning tasks, but it was less effective than tuning pre-trained language models (PLMs). 
Some papers focus on enhancing spatial reasoning by in-context learning. 
For instance,~\citealt{sharma2023exploring} proposed a prompting technique that decomposes the target spatial question into simpler questions and infers the answer to the final query based on the information gained from them.  
Another approach, proposed by ~\citealt{hu2023chainofsymbol}, replaced textual explanations in the Chain-of-Thought (CoT) by symbolic representations. 
While this method improved spatial understanding in simple environments, it was less applicable to diverse and complex spatial relations settings.
Lastly, some studies employ neuro-symbolic modeling that involves LLMs in the pipeline \citep{yang2023coupling, mirzaee2023disentangling}.
These approaches extract relations and obtain formal representations using LLMs. 
Then, the formal representation is passed to an external solver that includes engineered rules of reasoning to conduct logical reasoning and answer spatial questions.
While this method significantly improved multi-hop reasoning, it required formal representations at inference time, which posed a challenge. Furthermore, formalizing commonsense reasoning is difficult, if not impossible, for realistic domains.  

Unlike the previous studies, we exploit available spatial logical knowledge as a set of constraints in fine-tuning language models.
Compared to hard pipelines, this approach has the advantage that logical knowledge does not need to be complete and cover all possible situations.
Our approach utilizes synthetic data and logical representations during training only.
Numerous research studies have highlighted the utility of logical knowledge in solving question-answering tasks.
For instance, \citealt{prager-etal-2004-question} demonstrates that answering sets of questions jointly using their logical relationships enhances the performance in reasoning tasks.
\citealt{amizadeh2020neurosymbolic} directly used the logical formalization of the question as input to the model to improve the logical reasoning of vision and language models.
\citealt{sun2022reasoning} proposed framework with two main components. The first component is for encoding logical knowledge, and the second is for entailment of that knowledge.
Lastly, \citealt{kamali2024nesycoconeurosymbolicconceptcomposer} proposed a framework that integrates logical reasoning with generalizable grounding interpretation to enhance both grounding and reasoning in vision and language models. 

Utilizing logical knowledge has been proven effective in solving many other NLP problems. 
For example,~\citealt{lu2021neurologic} uses a heuristic function inspired by the A* algorithm to restrict the output of generative language models with a set of constraints, while \citealt{Qin2022COLDDE} uses a sampling method looking for the most optimal solution that satisfies the constraints.
Both of these techniques apply the constraints during inference, which differs from our proposed method for using them during the training.
Lastly, \citealt{faghihi2023gluecons} provided a benchmark to evaluate the benefits of logical constraints in several NLP problems under the DomiKnowS framework~\cite{faghihi2021domiknows}.
In this paper, we propose a novel approach for formulating spatial reasoning knowledge as logical constraints and exploit the existing techniques for the actual integration.
We utilize constraints solely during training, eliminating the need for access to logical constraints during inference where using them might be problematic for realistic domains and real-time applications.

\begin{figure*}[t]
    \centering
    \includegraphics[width=0.85\linewidth]{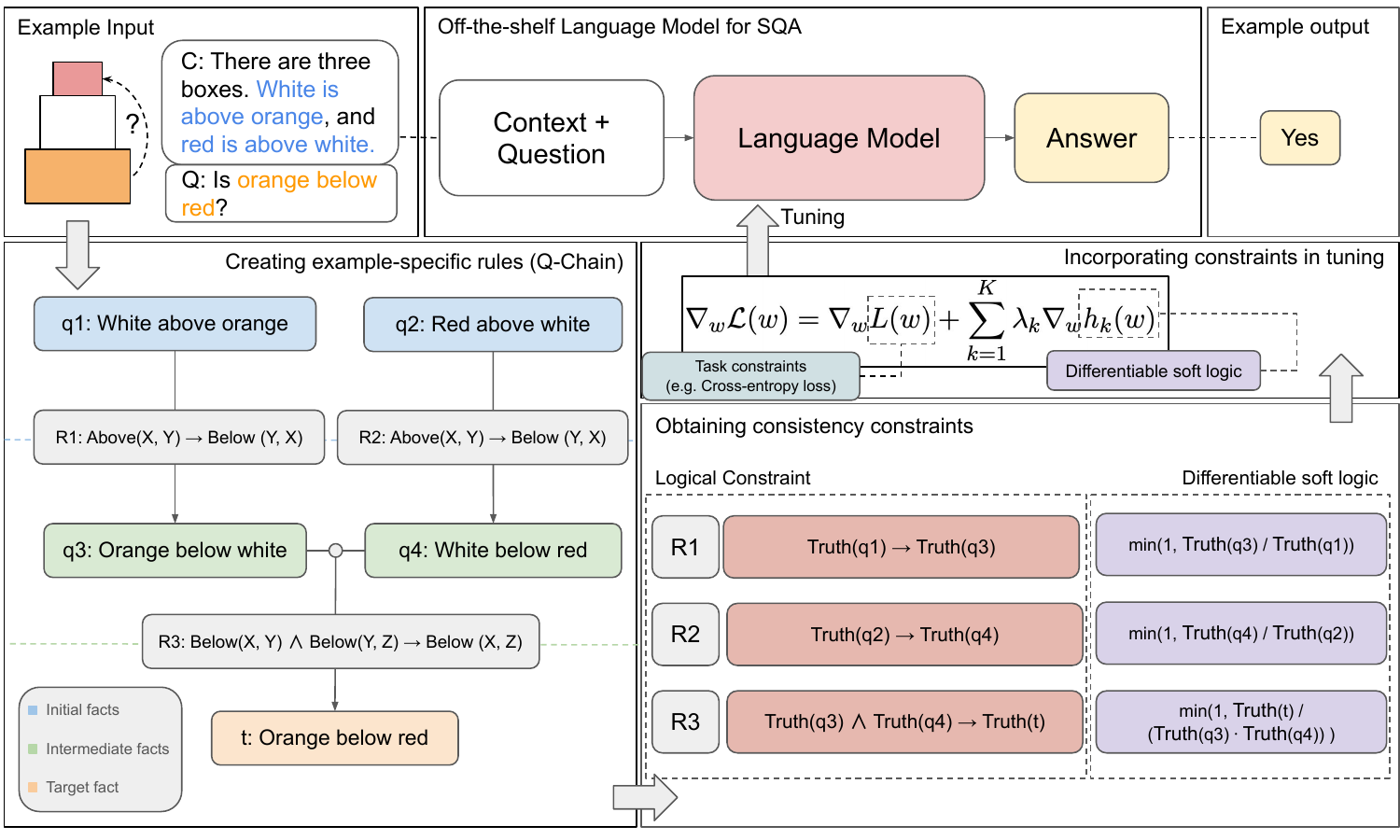}
    \caption{The pipeline for exploiting spatial logic in the training phase involves three steps: 1. Creating the chain of reasoning (Q-Chain) for each training example using annotated facts and rules from the training dataset. 2. Obtaining consistency constraints from spatial logical rules and answers inferred from questions generated using intermediate facts in Q-Chain. 3. Incorporating obtained consistency constraints into the loss function using differentiable soft logic during tuning. For more details, refer to Section~\ref{sec:creating_q_chain}.}
    \label{fig:chain}
\end{figure*}

\section{Methodology}
\subsection{Problem Definition}
For the spatial QA task, the input includes a textual context $C$ that describes a scene and a question $Q$ about the spatial relations between objects in the scene. 
The output is the answer(s) to the question, denoted as $A$. There are two types of questions/answers: Yes/No (YN) and find-relation (FR).
The answer domain is restricted to $A$ = $\{Yes, No\}$ for YN questions. 
The answers to FR questions depend on the dataset and the types of relations covered. 
For example, the answer domain for FR can be $A$ = \{$left$, $right$, $above$, $below$,\dots\}. 

\subsection{Backbone Language Models}
We use both encoder-based models from the BERT family~\cite{devlin2019bert} as well as generative models like Flan-T5~\cite{chung2022scaling} to assess the effectiveness of our proposed fine-tuning approach. 
We used other LLMs, such as GPT family and Llama3-8B, as comparison baselines without fine-tuning.

\subsubsection{Encoder-based Language Models}
\label{bert-arch}
We utilize BERT as the backbone architecture to consistently build on top of previous results in ~\citet{mirzaee2022transfer}. 
We fine-tune BERT by adding classification layers for QA tasks. 
We employ a binary classification layer for the YN domain while using multiple binary classification layers for the FR domain.
The motivation behind using multiple binary classification layers instead of one multi-label classification layer is that one question can have multiple labels as responses. 
The number of binary classification layers in the FR domain depends on the number of possible answers, $|A|$, in the problem setting. 
The input to the BERT model is constructed by concatenating the question, $Q$, separator token, $[SEP]$, and the context $C$. 
After feeding the input to BERT, we use the $[CLS]$ token from the last layer as the input to the classification layer(s) for final answer prediction.

\subsubsection{Generative Language Models}
\label{T5-arch}
We select Flan-T5~\cite{chung2022scaling} as the open-source generative baseline for fine-tuning. 
In the YN setting, processing the output is straightforward, as the first token can represent the response directly generated by the model. 
We compare the probability of the Yes and No tokens and select the one with the highest probability as the answer. 
However, this approach is not feasible in the FR setting due to the multi-label questions requiring multiple answers.
The input prompt to the Flan-T5 model is structured as follows: ``Answer based on the context: $C$ $Q$.'' 

For In-context Learning, we select Llama3-8B~\cite{llama3modelcard}, GPT3.5~\cite{brown2020language}, and GPT-4~\cite{openai2023GPT4} as the baseline LLMs using \textit{Zero-shot}, and \textit{Few-shot} settings.
\label{sec:in_context_normal}

\noindent\textbf{Zero-shot.} We prompt the LLMs to provide answers based on the scene description and the question without including any examples.

\noindent\textbf{Few-shot.} We randomly sample four pairs of questions and answers from the training dataset and concatenate them to the context and target question to form the few-shot setting.

\subsection{Training with Spatial Logic}
\label{sec:train_with_logic}
\label{sec:constraints_rule}
To address the challenge of multi-hop spatial reasoning, we employ symbolic knowledge during training.
We assume a formal knowledge base (KB) of rules, referred to as \textit{spatial logic}, is given during training time.
This spatial logic does not need to be exhaustive or cover all reasoning aspects; any available knowledge can be used to enhance the reasoning capabilities of the language model.
This work uses 79 reasoning rules taken from ~\citet{mirzaee2022transfer}. 
These rules are divided into four categories: converse, inverse, transitive, and transitive + topological. 
The KB covers rules between 15 spatial relations, including directional and topological relations, such as ``\textit{If x is above y, then y below x}.'' denoted formally as $Above(x, y) \Rightarrow Below(y, x)$.

Our main hypothesis is that supervision from high-level logical knowledge allows the model to capture more abstract patterns, thereby improving generalization to other domains.
To exploit the spatial logic in the training phase, we follow three steps:
\textbf{(1)} Creating the chain of reasoning for each training example. 
\textbf{(2)} Obtaining consistency constraints from the chain of reasoning. 
\textbf{(3)} Incorporating constraints in the loss function.
Please note that we will need the reasoning chain annotations for fine-tuning with spatial logic. SpaRTUN is the only synthetic dataset that provides such annotated formal representations, making it a suitable source of supervision for our tuning purpose.

\noindent\textbf{Creating example-specific rules.} \label{sec:creating_q_chain}
We exploit \textit{Spatial Logic} by automatically augmenting each training example with a chain of questions, denoted as $Q$-$Chain$. 
This process is illustrated in Figure~\ref{fig:chain}. 
We infer the answer to the final query from the input context using the \textit{logical resolution tree}, which outlines the logical steps for implication.
Our synthetic training data (i.e., SpaRTUN) provides logical representations of the initial facts expressed in the context.
We build the resolution tree using the \textit{forward chaining algorithm} to infer a specific target fact given a set of initial facts. 
In our example, the initial facts are  ``The white is above the orange'', denoted as $q_1$, and ``The red is above the white'', denoted as $q_2$. 
The target fact is ``The orange below the red,'' denoted as $t$.
The forward chaining approach is a reasoning method that starts with the initial facts in the knowledge base. It performs an exhaustive search for all applicable inference rules and applies them to the current facts to iteratively infer new ones until the target fact is reached or no new facts are inferred~\cite{forwardChainingDef}.
In the example of Figure~\ref{fig:chain}, we apply the spatial converse rules to the initial fact $q_1$ and $q_2$.
This approach yields the intermediate inferred facts, $q_3$, ``The orange is below the white,'' from $q_1$, and $q_4$, ``The white is below the red,'' from $q_2$. 
Finally, we search for the applicable spatial rule to combine $q_3$ and $q_4$, resulting in the target fact, $t$, ``The orange is below the red.''

\noindent\textbf{Obtaining consistency constraints.}
In this step, we obtain a set of consistency constraints based on the given resolution tree. We denote the truth value of a fact $q_i$ in the Q-Chain as $Truth(q_i)$, which will be True if the fact is True and otherwise is False. We create a consistency constraint for each step of the Q-Chain.
For example, in Figure~\ref{fig:chain}, 
step $R_1$ is applied to $q_1$ to infer $q_3$. This step implies that the $Truth(q_1)$ should be consistent with the $Truth(q_3)$. In other words, $q_1$,  ``white above orange,'' implies  $q_3$: ``orange below white,''.  The corresponding formal constraint for this example will be $Truth(q_1)\Rightarrow Truth(q_3)$.

\noindent\textbf{Incorporating Constraints.}
Our fine-tuning objective is to minimize the violation from the set of logical consistency constraints given for each training example. However, the logical expressions are not differentiable. To form a differentiable loss function, we use a soft logical surrogate of the constraints in the loss.
There are various types of conversions for mapping logical constraints into differentiable soft logic, including t-norm Product, t-norm \L ukasiewicz, and t-norm Gödel~\cite{li2019logicdriven}.
We use the t-norm Product to convert, which means, $\neg A$ becomes $1 - a$, $A$ $\land$ $B$ becomes $ab$, $A$ $\lor$ $B$ becomes $a + b - ab$, and $A \Rightarrow B$ becomes $min(1, \frac{b}{a})$, where $A$ and $B$ are concepts with probabilities $a$ and $b$ respectively. During the training, the probability of the truth of the facts ($Truth(q_i)$) involved in the constraints is obtained from the model prediction for computing the loss. 
We build on previous studies of optimizing an objective that combines the task performance and soft constraints~\cite{li2019logicdriven, asai2020logicguided,bamas2020primaldual} and form the following objective,
\begin{equation}
    \nabla_w \mathcal{L}(w) = \nabla_w L(w) + \sum_{k=1}^K \lambda_k \nabla_w h_k(w),
\end{equation}
where $L(w)$ is the task-performance loss function, i.e., Cross-Entropy loss, $K$ is the number of logical constraints related to the current $Q$-$Chain$, $h_k$ measures the violation from a given logical constraint in soft logic differentiable form, and $\lambda_k$ is a learning parameter indicating the importance of obeying $k^{th}$ rule. 
We set the $\lambda_k$ to 1 for all constraints to ensure equal importance.

In fact, in this work, we utilize the implementation in the DomiKnowS framework~\cite{faghihi2021domiknows} to integrate the logical constraints. 
This framework provides a declarative language for incorporating symbolic knowledge as logical constraints. 
We present the original logical forms in the program, and DomiKnowS automatically converts them into a differentiable form for inclusion in the loss function.
We employ an optimization technique referred to as the Primal-Dual (PD) program that uses a dual formulation of the original objective~\cite{bamas2020primaldual} for fine-tuning.
Programming for building a model in DomiKnowS requires defining a graph of concepts and relationships, along with logical rules between concepts. The program for our problem setting is provided in Appendix~\ref{appendix:domiknows}.

\subsection{In-context Learning with Spatial Logic}
\label{section:in_context_learning}

We design a set of experiments to analyze and compare the LLMs' performance against our proposed fine-tuning method. We utilize spatial logical reasoning to manually create the rationale of few-shot examples for our in-context learning settings, including Chain-of-Thoughts~\cite{wei2023chainofthought}, Logical Representation, and Chain-of-Symbols~\cite{hu2023chainofsymbol}.

\noindent\textbf{Chain-of-Thought (CoT).} To enable CoT, we provide instructions that require an explanation before answering the question. 
For each example response, we include a manually crafted rationale based on spatial logical rules. We provide these examples with instructions and target questions as input to the model to generate the answer. 
Multiple examples of this CoT setting are demonstrated in Appendix~\ref{appendix:CoTExample}.

\noindent\textbf{Logical Representation (LR).} 
In this setting, we follow the same prompt structure as CoT but use logical representations instead of natural language.
We obtain the first order logical representations (predicate-argument structures) by manually converting the CoT explanations. 
The predicate-argument structure $R(obj_i, obj_j)$ represents the relation $R$ between $obj_i$ and $obj_j$, where $obj_i$ and $obj_j$ denote object $i$ and object $j$, respectively.
An example of the LR counterpart of the CoT phrase is in Table~\ref{tab:CoT_conversion}.

\noindent\textbf{Chain of Symbol (CoS).} This symbolic representation was introduced by~\citet{hu2023chainofsymbol}. 
We explicitly follow the proposed process in~\citet{hu2023chainofsymbol} for converting spatial expressions from natural language into symbolic representation.
An example of the CoS counterpart of CoT expression is shown in Table~\ref{tab:CoT_conversion}.

\begin{table}[h!]
    \begin{adjustbox}{width=\columnwidth - 2mm, center}
    \begin{tabular}{| c | c |}
        \hline
        Domain & Spatial context\\
         \hline
         CoT & large red square is to the left of a small green square\\
         \hline
        LR & Left(large red square, small green square)\\
         \hline
        CoS & (large, red, square) $<$ (small, green, square)\\
         \hline
    \end{tabular}
    \end{adjustbox}
    \caption{An example of a spatial relation used in Chain of Thought (CoT), Logical Representation (LR), and Chain of Symbol (CoS).}
    \label{tab:CoT_conversion}
\end{table}

\noindent\textbf{Step-by-Step Prompting.}
\label{sec:stepbysetpReasoning}
From the SpartQA-Human dataset, we observed that scene descriptions are often complex and lengthy. 
To deal with this issue, we introduce an additional prompting step using LLMs to break down the context before answering the question. 
Each generated line of context contains a simple spatial description.
For example, given the original context, ``\textit{There exists a big red square and a medium red square in a block called A. The big one is touching the right edge of the block.'' }
the expected context generated by the LLM would be \textit{``The big red square in block A. <newline> The big red square is touching the right edge of block A...}''.
More examples are provided in Appendix~\ref{appendix:in_context_example}. 
We refer to this prompting technique as \textit{Step-by-Step}, which can be applied before any ICL method.


\section{Experiments}

\subsection{Datasets}

Our experiments are conducted on all existing datasets, including SpaRTUN, SpartQA-Human, ReSQ, and StepGame. 

\noindent\textbf{SpaRTUN}~\cite{mirzaee2022transfer} is a synthetic SQA dataset. This dataset contains both YN and FR types of questions that require multi-hop reasoning to answer. It covers a wide range of spatial relations.\footnote{The dataset also includes Don't Know questions, but for simplicity, we apply the closed-world assumption and label those questions as No.}
The answer for YN is $A$ = $\{Yes,No\}$,  
while the answer for the FR is a subset of 15 spatial relations, e.g., $\{left, behind\}$, and $\{above, inside\}$.
The dataset provides the chain of reasoning annotations that we use to create the $Q$-$Chain$s for our proposed training method. 
This dataset is used only for training because it is synthetic and the baselines performance reported in~\cite{mirzaee2022transfer} on its test split is already high.

\noindent\textbf{SpartQA-Human (Ver.1)}~\cite{mirzaee2021SpartQA} is a small human-annotated SQA dataset. The dataset contains both YN and FR types of questions. 
We only use the YN question types for fine-tuning and evaluation. The YN questions require more reasoning steps and are shown to be more challenging based on previous research results in~\citet{mirzaee2022transfer}.

\noindent\textbf{SpartQA-Human (Ver.2)} We follow the methodologies from~\cite{mirzaee2021SpartQA} to increase the size of the human annotated dataset. 
The objective is to enrich the SpartQA-Human with a more extensive test using diverse patterns and questions.
Annotators included authors and an undergraduate student who was paid as a research assistant. We extended both types of YN and FR questions.

\noindent\textbf{ReSQ}~\cite{10.1007/978-3-319-65813-1_32, mirzaee2022transfer} is a small realistic SQA dataset with multi-hop reasoning. The depth of reasoning is smaller than other synthetic datasets, but it usually requires commonsense knowledge to infer the answer. 
The annotations include the depth of reasoning denoted as $k$, which can be 1 or 2. 
The questions that required complex reasoning and commonsense knowledge are annotated as \textit{unclassified} depth. 
12.30\% of the test examples fall under the $k$=1 split, 23.93\% fall under the $k$=2 split, and the rest are unclassified.
This dataset only has YN questions. We use this dataset to evaluate the performance of our models in realistic domains.

\noindent\textbf{StepGame}~\cite{shi2022StepGame} is a synthetic SQA dataset that contains a diverse set of multi-hop reasoning questions involving spatial relations, with annotated reasoning depth denoted as $k$ .
The questions in this dataset include up to 10 hops of reasoning, allowing for a comprehensive evaluation of  models' reasoning capability at varying depths.
This dataset contains only FR questions. The domain of answer is $a$ = \{$left$, $right$, $above$, $below$, $overlap$, $lower$-$left$, $lower$-$right$, $upper$-$left$, $upper$-$right$\}. 

\subsection{Experimental Models for Tuning}


\begin{table*}[ht!]
    \begin{adjustbox}{width=\columnwidth + \columnwidth - 60mm, center}
    \begin{tabular}{ | l | c | c | ccc | c| c |} 
     \hline
       & \multicolumn{2}{|c|}{SpartQA-Human} & \multicolumn{4}{|c|}{ReSQ} & \\
       \hline
        Model & Ver.$1$ & Ver.$2$ &  $k$=1 & $k$=2& unclassified & All & \# \\
     \hline
      BERT   & $54.54$ & $53.57$ & $70.67$ & $ 56.85$ & $60.66$ & $60.98$ & 1 \\ 
     BERT-T   & $55.94$& $58.03$ & $\mathbf{76.00}$ & $54.79$ & $\mathbf{61.18}$ & $61.15$& 2 \\
     BERT-T+$Q$-$Chain$ (Our) & $\mathbf{59.44}$ & $\mathbf{58.92}$  & $72.00$ &  $\mathbf{58.90}$& $59.90$ & $\mathbf{61.31}$ & 3\\

     \hline
     Flan-T5   & $54.54$ & $60.71$  & $74.67$ & $56.16$ & $61.44$ & $61.80$ & 4\\ 
     Flan-T5-T   & $49.65$ & $57.14$  & $81.33$ & $54.79$ & $61.44$ & $62.30$ & 5\\
     Flan-T5-T+$Q$-$Chain$ (Our) &  $\mathbf{55.94}$ & $\mathbf{61.61}$  & $\mathbf{81.33}$ & $\mathbf{57.53}$& $\mathbf{63.75}$ & $\mathbf{64.43}$ & 6 \\
     \hline
     GPT3.5 (zero-shot)  & $58.04$ & $58.03$ & $74.67$ & $60.95$ & $66.58$  &  $66.22$ & 7 \\ 
     GPT3.5 (few-shot) & $62.23$  & $58.92$ & $84.00$ & $68.49$ & $68.12$ &  $70.16$ & 8\\ 
     GPT3.5 (LR) & $64.33$ & $59.82$ & - & - & - & - & 9 \\ 
     GPT3.5 (CoS) & $60.14$ & $64.28$ & - & - & - & - & 10 \\ 
     GPT3.5 (CoT) & $65.73$ & ${71.43}$ & $\mathbf{86.67}$ & $67.12$ & $68.64$ &  $70.49$ & 11 \\ 
     GPT-4 (zero-shot) & $\mathbf{77.62}$ & $68.75$ & $84.00$ & $73.97$ & $\mathbf{76.86}$ &  $\mathbf{77.05}$& 12 \\
     Llama-3 (zero-shot) & $61.54$  &$50.89$& $80.00$ & $64.38$ & $67.35$ & $68.20$ & 13\\
     Llama-3 (few-shot) & $62.94$ & $60.71$& $82.67$ & $69.86$ & $71.46$ & $72.46$ & 14\\
     Llama-3 (LR) & $60.14$ & $\mathbf{72.32}$ & - & - & - & - & 15 \\ 
     Llama-3 (CoS) & $67.13$ & $61.60$ & - & - & - & - & 16 \\ 
     Llama-3 (CoT) & $67.83$ &$70.54$& $82.76$ & $\mathbf{76.03}$ & $67.10$ & $71.15$ & 17\\
     \hline
     PistaQ* & $75.52$ & - & - & - & - & $47.21$ & 18\\
     \hline
    \end{tabular}
    \end{adjustbox}
    \caption{Accuracy of SpartQA-Human and ReSQ with various models. For ReSQ, $k$ is the number of the reasoning steps required to answer the questions. \textit{Unclassified} indicates the cases in which $k$ was challenging for human annotators to determine. The results from PistaQ* reported from \citet{mirzaee2023disentangling}.}
    \label{tab:ReSQ_with_k}
\end{table*}

\noindent\textbf{BERT-family.} We use \textit{bert-based-uncased} as the initial checkpoint, referred to as BERT. This model is only fine-tuned with the target dataset in each experiment. In some of the initial experiments we used RoBERTa as well. However, due to insignificance of the differences, we continued other model variations with BERT. The results of RoBERTa can be found in Appendix~\ref{appendix:roberta}.

\noindent\textbf{BERT-T.}  We fine-tune the model with SpaRTUN first. Then, we further fine-tune with the target dataset of each experiment.

\noindent\textbf{BERT-T + $\mathbf{Q}$-$\mathbf{Chain}$.}  We adhere to the methodology explained in Section~\ref{sec:train_with_logic} to integrate logical knowledge into the BERT-T model via fine-tuning with constraints using the DomiKnowS framework~\cite{faghihi2021domiknows}. 

\noindent\textbf{Flan-T5.} We follow the methodology described in Section~\ref{T5-arch}. We selected the \textit{flan-t5-base} in this experiment as the starting checkpoint. To reduce computational costs by decreasing the number of training parameters, we employ the LoRA adapter~\cite{hu2021lora}.

\noindent\textbf{Flan-T5-T.} We utilize the same setting as the Flan-T5 model. We fine-tune the model first with SpaRTUN and then with the target dataset.

\noindent\textbf{Flan-T5-T + $\mathbf{Q}$-$\mathbf{Chain}$.} We integrate the logical knowledge into Flan-T5-T model using proposed method in section~\ref{sec:train_with_logic}. The backbone is the same as the baseline model without any changes. 

\noindent\textbf{Experimental Setup.} We used the same hyperparameters across all experiments. The learning rate was selected from \{$1e^{-5}$, $8e^{-6}$, $1e^{-6}$\}.
For SpartQA-Human and ReSQ, the number of epochs was 100, while for StepGame, it was 30. 
For fine-tuning with SpaRTUN, we used 12 epochs for BERT and 8 epochs for Flan-T5. 
The final hyperparameters were chosen using the development set of the target dataset.
In all experiments, we used Cross-Entropy Loss as the loss function and Adam as the optimizer, with the \textit{amsgrad} parameter set to \textit{True}. 
The experiments were run on 8 A6000 GPUs, requiring approximately 100 GPU hours.

\subsubsection{Experimental Models for In-context} 

We evaluate the performance of three selected LLMs for in-context learning setting including GPT3.5, GPT-4, and Llama3-8B, using ReSQ and SpartQA-Human to compare with our fine-tuning approach. 
The selected examples and prompts for each in-context learning method can be found in the Appendix~\ref{appendix:in_context_example}.

\noindent \textbf{Zero-shot}. We directly ask LLMs to answer the question given the corresponding context.

\noindent\textbf{Few-shot}. We provide LLMs with four randomly selected examples from the training data. 

\noindent\textbf{CoT}. We augment the \textit{few-shot} setting with reasoning explanations along with the \textit{CoT} examples. 

\noindent\textbf{LR}. We provide the few-shot examples with the reasoning explanation in the logical format described in Section~\ref{section:in_context_learning}.

\noindent\textbf{CoS}. We further alter the \textit{few-shot} examples with the CoS symbolic forms explained in Section~\ref{section:in_context_learning}.

\begin{table*}[ht!]
    \begin{adjustbox}{width=\columnwidth + 50mm, center}
    \begin{tabular}{ |l|c c c c c c c c c c| } 
     \hline
     Model & $k$=1 & $k$=2 & $k$=3 & $k$=4 & $k$=5 & $k$=6 & $k$=7 & $k$=8& $k$=9 & $k$=10 \\
     \hline
    BERT & $98.51$ & $95.53$ & $91.68$ & $66.71$ & $49.11$ & $41.47$ & $41.47$ & $32.09$ & $28.94$ & $28.16$ \\ 
    BERT-T & $98.50$ & $95.32$ & $\mathbf{93.26}$ & $\mathbf{76.78}$ & $\mathbf{66.36}$ & $58.76$ & $53.70$ & $46.27$ & $42.71$ & $40.12$ \\ 
    BERT-T+$Q$-$Chain$ (Our) & $\mathbf{98.70}$ & $\mathbf{96.45}$ & $93.03$ & $74.58$ & $64.95$ & $\mathbf{59.04}$ & $\mathbf{54.38}$ & $\mathbf{49.23}$ & $\mathbf{45.36}$ & $\mathbf{44.05}$ \\
    \hline
    Llama-3 (few-shot) & $38.01$ & $27.87$ & $24.15$ & $21.27$ & $19.75$ & $18.03$ & $16.88$ & $15.52$ & $15.17$ & $14.70$ \\ 
    \hline
    GPT3 (few-shot) & $55.00$ & $37.00$ & $25.00$ & $30.00$ & $32.00$ & $29.00$ & $21.00$ & $22.00$ & $34.00$ & $31.00$ \\ 
    GPT3 (CoT) & $61.00$ & $45.00$ & $30.00$ & $35.00$ & $35.00$ & $27.00$ & $22.00$ & $24.00$ & $23.00$ & $25.00$ \\ 
    GPT3(d2) + ASP & $92.60$ & $89.90$ & $89.10$ & $93.80$ & $92.90$ & $91.60$ & $91.20$ & $90.40$ & $89.00$ & $88.30$ \\
    \hline

     \hline
    \end{tabular}
    \end{adjustbox}
  \caption{Accuracy of StepGame on several models including results of GPT3 reported from~\citealt{yang2023coupling}.}
  \label{tab:StepGame_result}
\end{table*}

\subsection{Results and Discussion}
\begin{figure}[t]
    \centering
    \includegraphics[width=\linewidth]{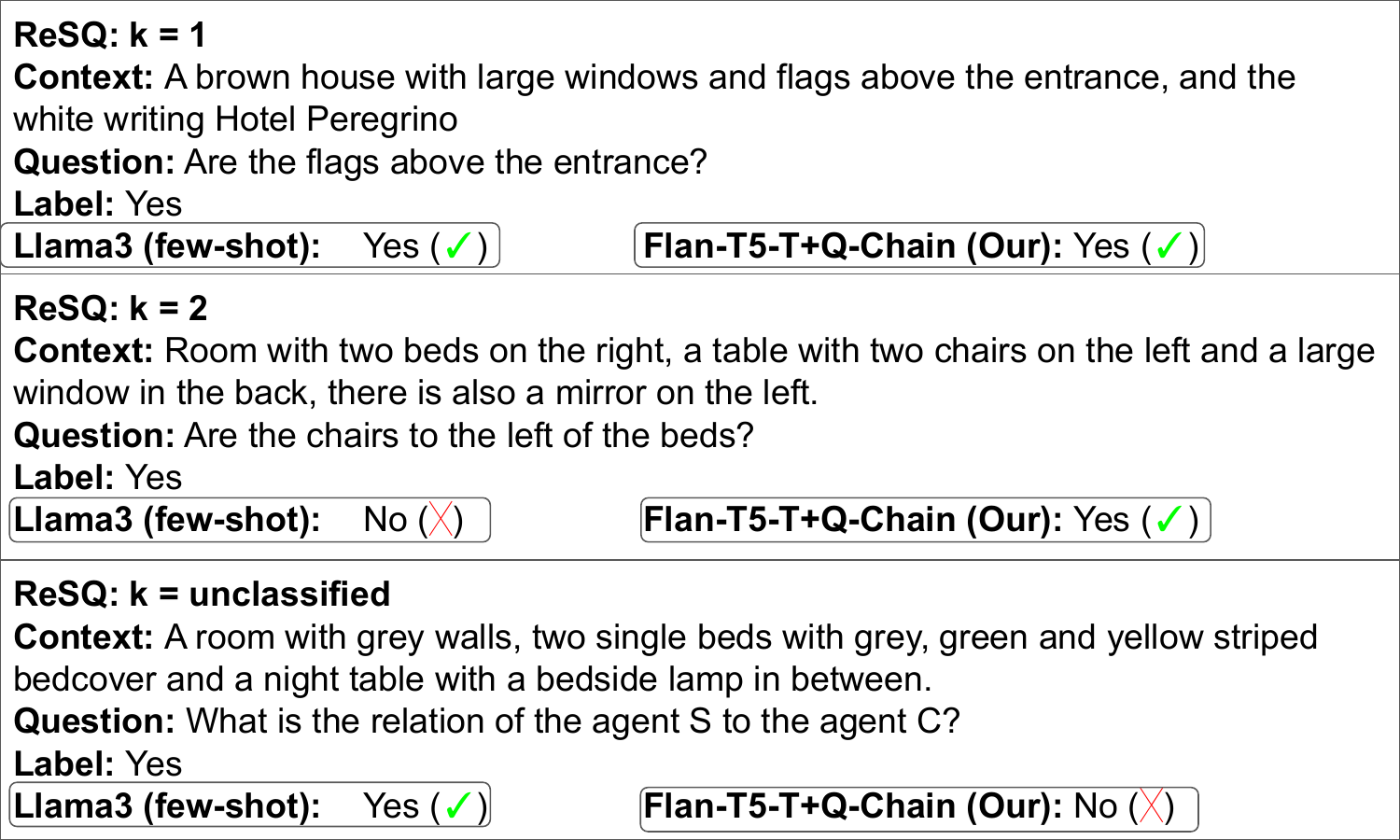}
    \caption{Three examples of each category in the ResQ dataset with responses from Llama3 (Few-shot) and Flan-T5 + Q-Chain (Ours).}
    \label{fig:resq_error_ex}
\end{figure}

\noindent\textbf{Realistic Domain.} 
ReSQ serves as a realistic SQA domain. 
Table~\ref{tab:ReSQ_with_k} shows that using the $Q$-$chain$ is effective for both BERT and Flan-T5, with notable improvement on Flan-T5.
To better understand these results, we analyzed the performance across different splits of ReSQ based on the manually annotated depth of reasoning required to answer the questions.
The first two splits include questions requiring one or two hops of reasoning, denoted as $k=1$ and $k=2$. 
The \textit{unclassified} split covers questions where reasoning depth is hard to determine and rely more on commonsense knowledge. 
An example of each case is provided in Figure~\ref{fig:resq_error_ex}.

Our observation from Table~\ref{tab:ReSQ_with_k} reveals that our model consistently improves performance on $k=2$ but negatively impacts BERT on $k=1$ and the \textit{unclassified} categories.
This suggests that our fine-tuning approach is more effective when deeper reasoning is needed but is less effective for questions requiring commonsense knowledge.
On the other hand, LLMs outperform fine-tuning approaches on ReSQ, specifically in the \textit{unclassified} subset. 
They consistently show $2$-$13$\% higher performance than Flan-T5+T+$Q$-$chain$ (lines 7-13).
These results indicate that one of LLMs' superior performances is due to their commonsense knowledge while our approach performs better in multi-hop reasoning. 
Nevertheless, we observe that logic-based fine-tuning enhances Flan-T5 more significantly than BERT on the \textit{unclassified} subset. 
This suggests that the $Q$-$chain$ can facilitate complex reasoning, particularly when applied to a model with stronger commonsense capabilities.
Note that applying LR and CoS is not straightforward on ReSQ due to the challenges associated with extracting logical representations from realistic domains text. 
This challenge is evident in the results reported for PistaQ*, which employs a similar approach. PistaQ* achieved only $47.21$\% accuracy on ReSQ, despite significant improvements observed in synthetic domains such as SpartQA-Human.

\begin{table}[t!]
    \centering
    \small
    \begin{tabular}{ |l| c | c| } 
     \hline
     Model & Raw Context & Step by step\\
     \hline
     GPT3.5 (zero-shot) & $58.04$ & $63.64$ \\ 
     GPT3.5 (few-shot) & $62.23$ & $64.33$  \\ 
     GPT3.5 (CoT) & $65.73$ & $67.83$ \\ 
     GPT3.5 (LR) & $64.33$ & $59.44$ \\ 
     GPT3.5 (CoS) & $60.14$ & $58.74$ \\ 
     GPT-4 (zero-shot) & $\mathbf{77.62}$ & $\mathbf{78.32}$ \\
     \hline
    \end{tabular}
    \caption{The accuracy of LLMs on SpartQA-Human.}
    \label{tab:GPT_human_result}
\end{table}

\begin{figure}[t]
    \centering
    \includegraphics[width=\linewidth]{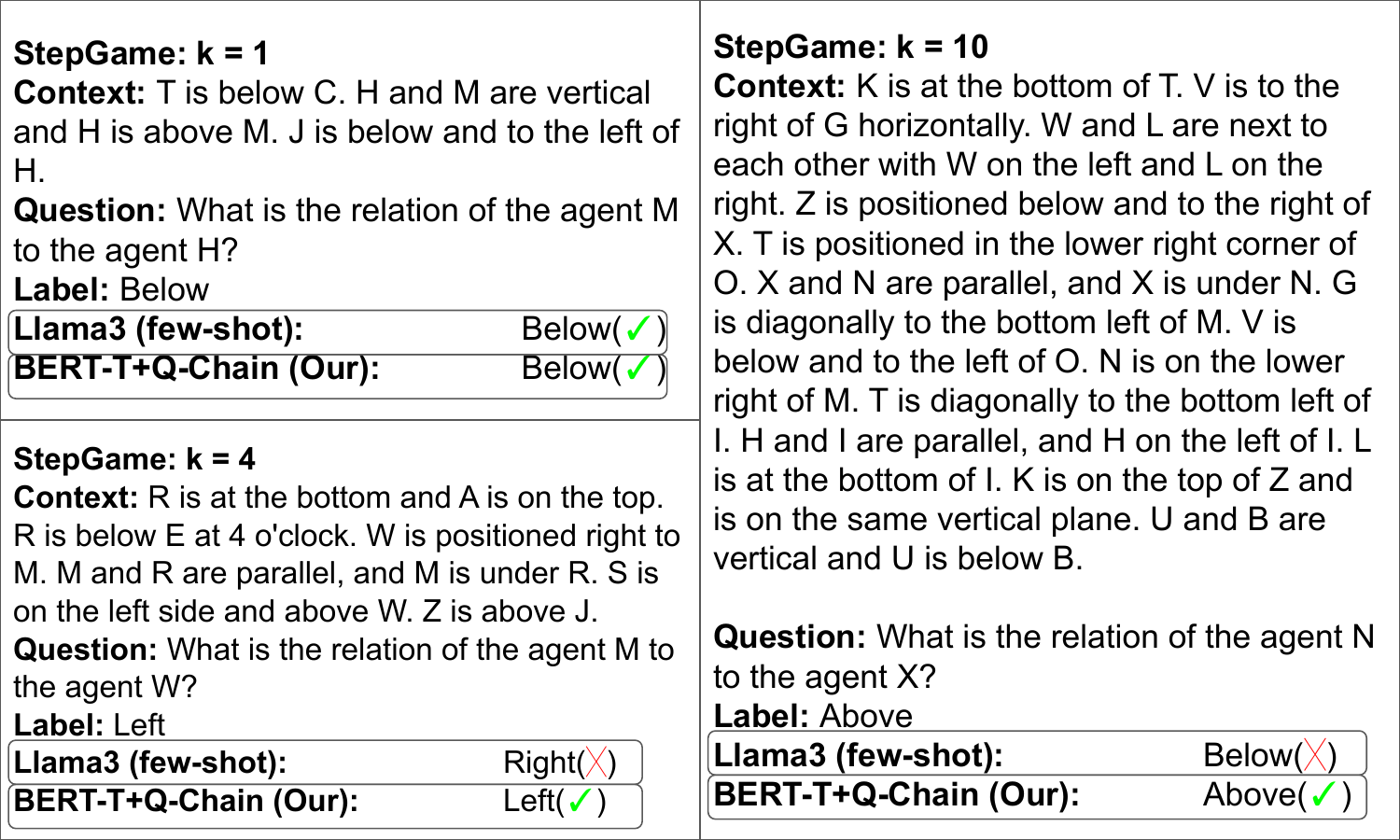}
    \caption{Three examples of each category in the StepGame dataset with responses from Llama3 (Few-shot) and BERT-T + Q-Chain (Our).}
    \label{fig:stepgame_error_ex}
\end{figure}

\noindent\textbf{Synthetic Domain with More Complex Logical Reasoning.}
SpartQA-Human and StepGame are synthetic domains used in our experiments. 
According to Table~\ref{tab:ReSQ_with_k}, we consistently observe improvements with our proposed $Q$-$chain$ in the synthetic domain which typically requires considerable reasoning hops.
Moreover, the gap between small PLMs and LLMs is much smaller in this domain compared to the realistic domain (ReSQ).
This is expected, as we concluded in the results of the realistic domain in Table~\ref{tab:ReSQ_with_k}, that LLMs are better at commonsense reasoning than complex multi-hop spatial reasoning.
The hypothesis is further supported when assessing the proposed method on StepGame. 
As can be observed in Table~\ref{tab:StepGame_result}, the fine-tuning method consistently demonstrates significant positive differences in all reasoning steps compared to LLMs.
The struggle of GPT3 on reasoning over StepGame is also investigated in~\cite{yang2023coupling}. 
Our proposed method consistently improves by $1$-$4$\% on a higher number of reasoning hops ($k$=$6$ to $k$=$10$), similar to the observation results of ReSQ. 
These results confirm our primary hypothesis that our proposed method equips the models with a higher level of logical abstraction to conduct more complex logical reasoning.
We provide three examples of responses from this dataset to qualitatively demonstrate the improvements in Figure~\ref{fig:stepgame_error_ex}. 
We also included results from GPT-3(d2) + ASP and PistaQ*, both of which demonstrate significant effectiveness in multi-hop reasoning. 
These models perform relation extraction to obtain a formal logical representation using LLMs, while reasoning is handled separately using external tools with engineered rules of reasoning. 
However, these pipelines face two key limitations. 
First, they require formal representations at test time, effective only on synthetic data. 
Second, they do not fine-tune the language models for better reasoning, relying instead on external tools. 
This approach contrasts with ours, which focuses on directly improving the language models' reasoning capabilities.

\noindent\textbf{In-context Learning.} 
This section focuses only on comparing the variations of in-context learning for spatial reasoning. As shown in Table~\ref{tab:ReSQ_with_k} and Table~\ref{tab:GPT_human_result}, CoT proved to be the most effective approach for spatial reasoning in both GPT and Llama. 
We also evaluated the impact of integrating formal representations in in-context learning, referred to as the LR approach. 
As evident in Table~\ref{tab:GPT_human_result} (line 4), using formal representations marginally degraded performance. 
In most cases, utilizing natural language in the CoT proved to be more effective.
Additionally, we evaluate CoS symbolic representations, finding that CoT and LR outperformed CoS using the same examples.
This highlights the advantage of natural language reasoning over symbolic reasoning for in-context learning. 
We further applied step-by-step prompting explained in Section~\ref{sec:stepbysetpReasoning} to evaluate more sophisticated prompting techniques on LLMs.
Table~\ref{tab:GPT_human_result} shows that this technique achieves $2$-$5$\% better results than zero-shot, few-shot, and CoT using raw context. 
The results demonstrate that more straightforward and shorter contexts improve LLMs' spatial understanding. This is consistent with previous results on other domains and tasks~\cite{li-etal-2023-compressing, liu-etal-2024-lost}.
Lastly, we compare open-source Llama and closed-source GPT models. 
Our results indicate that Llama-3 is competitive with GPT-3.5, showing only minor differences in the YN domain as shown in Table~\ref{tab:ReSQ_with_k}. 
Thus, we plan to prioritize open-source models in future research.

\section{Conclusion}
Given the importance of spatial reasoning in many real-world applications, we focus on improving this reasoning skill in language models. 
We equip LLMs with neuro-symbolic supervision to leverage logical knowledge during fine-tuning. This practice boosts their capability of capturing the necessary level of abstraction for spatial reasoning and enhances the generalizability for transferring knowledge across domains.
Our constraint-based training technique achieves higher accuracy than other competitive Spatial Question-answering baselines across various benchmarks. Furthermore, the results indicate that our model performs better in scenarios requiring more reasoning steps. Lastly, we compare our models with state-of-the-art LLMs. Experimental comparisons show that while large LLMs like GPT3.5 perform better in commonsense reasoning, we achieve higher performance in multi-hop spatial question-answering with tuning comparatively smaller language models like BERT with spatial logic.

\section*{Limitations}
While we improve the reasoning capability of the models, our approach by no means solves the general reasoning problem. The trustworthiness and reliability of the LLMs are still a research challenge. Our models will need GPU resources to run, which might be expensive. Our research is based on open-source data and code, and the results will be reproducible. We do not see any ethical concerns in our research approach or the impact of our results. Our data, being limited to the spatial reasoning domain, does not include any specific type of bias that can harm minority people belonging to any particular gender or race. 

\section*{Acknowledgment}
This project is partially supported by the Office of Naval Research (ONR) grant N00014-23-1-2417. Any opinions, findings, and conclusions or recommendations expressed in this material are those of the authors and do not necessarily reflect the views of Office of Naval Research. 
We thank anonymous reviewers for their constructive feedback, which greatly helped us improve this manuscript.


\bibliography{latex/ref}

\appendix

\section{Data Statistic}
\label{appendix_data_stat}

The data statistic of each dataset is shown in Table~\ref{tab:data_statistic}.
\begin{table}[ht]
    \begin{adjustbox}{width=\columnwidth -0mm, center}
    \begin{tabular}{ |c|c c c| } 
     \hline
     Dataset & Train & Test & Dev  \\
     \hline
     SpaRTUN (YN) & 20633 & 3232 & 3152\\ 
     SpaRTUN (FR) & 18400 & 2818 & 2830\\ 
     $Q$-$Chain$ (YN) & 56987 & - & -\\ 
     $Q$-$Chain$ (FR) & 46750 & - & -\\ 
     \hline
     SpartQA-Human(Ver.1) & 161 & 143 & 51\\ 
     SpartQA-Human(Ver.2) & 200 & 112 & 60\\ 
     \hline
     ReSQ & 1008 & 610 & 333 \\
     \hline
     StepGame & 50000 & 5000 & 100000 \\
     \hline
    \end{tabular}
    \end{adjustbox}
    \caption{Size of each SQA benchmark used in experiments}
    \label{tab:data_statistic}
\end{table}

\section{RoBERTa results}\label{appendix:roberta}
We provide results from the same setting as the BERT using RoBERTa in Table~\ref{appenTableResult} and found that the overall results are slightly lower than BERT for typical transfer learning. So, we decided to use only BERT for the remaining experiments.

\begin{table*}[ht!]
    \small
    \centering
    \begin{tabular}{ | l | c | c | ccc | c| c |} 
     \hline
       & \multicolumn{2}{|c|}{SpartQA-Human} & \multicolumn{4}{|c|}{ReSQ} & \\
       \hline
        Model & Ver.$1$ & Ver.$2$ &  $k$=1 & $k$=2& unclassified & All & \# \\
     \hline
      BERT   & $54.54$ & $53.57$ & $70.67$ & $ 56.85$ & $60.66$ & $60.98$ & 1 \\ 
     RoBERTa   & $54.54$ & - &  $76.00$ & $56.16$ & $58.87$ & $60.33$ & 2\\ 
     BERT-T   & $55.94$& $58.03$ & ${76.00}$ & $54.79$ & ${61.18}$ & $61.15$& 2 \\
     RoBERTa-T   & $49.65$ & - & $64.67$ & $57.87$ & $55.78$ & $56.72$ & 4\\
     BERT-T+$Q$-$Chain$ (Our) & ${59.44}$ & ${58.92}$  & $72.00$ &  ${58.90}$& $59.90$ & ${61.31}$ & 3\\
     \hline
    \end{tabular}
    \caption{Accuracy of SpartQA-Human and ReSQ with BERT and RoBERTa for comparison between these models.}
    \label{appenTableResult}
\end{table*}

\section{In-context Learning Examples}
\label{appendix:in_context_example}

The example of each in-context learning prompt and example is indicated below.

\subsection{Few-Shot}

$\mathbf{system:}$ You will be given a story and question. Then, you will answer either only Yes or No based on given story. Candidate answer: [Yes, No]

$\mathbf{user:}$ There exist a big red square, a big red triangle, a medium red square, and a small green circle in a block called A. The triangle and medium square are touching the bottom edge of the block. The big and medium square are touching the right edge of the block. And the circle is above the big square which is above the medium square. There is another block called B to the left of block A. A medium green square is touching block B's left edge and below a medium red square. The medium red square is above and to the left of a small green square. Also a medium red triangle is below and to the left of the small square. Is the medium red triangle below the red square in block B?

$\mathbf{assistant:}$ Yes

$\mathbf{user:}$ There are three blocks A, B and C. Block A has a medium white triangle and a small red rectangle. There exists a medium white rectangle to the left of the triangle and to the right of the red rectangle. Aove block A there is block B. It has a small white oval touching the bottom edge of the block and a medium white triangle to the right of it. Below block A there is block C. Its has a small green triangle touching the top edge of the block. A medium white triangle is below a small green rectangle. Another small green triangle is below and to the right of the white triangle. Are all rectangles above a white triangle?'

$\mathbf{assistant:}$ Yes

$\mathbf{user:}$ There is a block called A. Inside A, a black triangle above and to the left of a black circle. There is another block called B which is to the right of A. B has two yellow circles, a blue circle, and a blue square. The blue circle is touching the left edge of B. One of the yellow circles is to the right of the blue circle and to the left of the blue square. The blue square is touching the right edge of B and above the other yellow circle, Which also touches the right edge of B. Does block B have all of the circles inside it?

$\mathbf{assistant:}$ No

$\mathbf{user:}$ There is a block called A. A has a large blue square near and below a medium blue triangle. Far to the left of the medium blue triangle is a small yellow square. Far above the medium blue triangle is a small blue triangle. There is another block called B, which is to the right of A. In B, There is a small yellow circle touching the left edge of B. Far to the right of the circle is a large yellow square. Near and above the large square is a small yellow square. There is another block called C, which is to the right of B. In C, there is a small blue circle near and to the left of a large yellow triangle. The large yellow triangle is near and to the left of a large blue triangle. The large blue triangle is far above a large blue circle. The large blue circle is touching the bottom edge of C. Are the large blue things in C to the left of the large blue thing in A?

$\mathbf{assistant:}$ No

$\mathbf{user:}$ There is a block called A. In A there is a red triangle Which is above a black circle. The black circle is touching the bottom of A. A yellow square is touching the top edge of A, and below that, a red oval is touching the right edge of A. There is a second block called B, to the left of A. Inside B are 3 squares: a brown, a green, and a red. The green square is above the red square and touching the left edge of B. The brown square is touching both the bottom edge and the right edge of B. The red square is touching the left edge of the brown square. Is the oval to the right of the brown thing?

\subsection{Chain of Thought}\label{appendix:CoTExample}

$\mathbf{system:}$ You will be given story and question. Then, you will answer either Yes or No with explanation based on given story step by step. Candidate answer: [Yes, No]

$\mathbf{user:}$ There exist a big red square, a big red triangle, a medium red square, and a small green circle in a block called A. The triangle and medium square are touching the bottom edge of the block. The big and medium square are touching the right edge of the block. And the circle is above the big square which is above the medium square. There is another block called B to the left of block A. A medium green square is touching block B's left edge and below a medium red square. The medium red square is above and to the left of a small green square. Also a medium red triangle is below and to the left of the small square. Is the medium red triangle below the red square in block B?

$\mathbf{assistant:}$ From information provided:The medium red square is above a small green square in B.Therefore, the small green square is below the medium red square in B. The medium red triangle is below the small square in B, and the small green square is below the medium red square in B.Therefore, the medium red triangle is below the medium red square in B. 
Answer: Yes

$\mathbf{user:}$ There are three blocks A, B and C. Block A has a medium white triangle and a small red rectangle. There exists a medium white rectangle to the left of the triangle and to the right of the red rectangle. Aove block A there is block B. It has a small white oval touching the bottom edge of the block and a medium white triangle to the right of it. Below block A there is block C. Its has a small green triangle touching the top edge of the block. A medium white triangle is below a small green rectangle. Another small green triangle is below and to the right of the white triangle. Are all rectangles above a white triangle?'

$\mathbf{assistant:}$ From given information: A medium white triangle is below a small green rectangle in C.Then, small green rectangle is above medium white triangle in C.A small red rectangle in block A, a medium white triangle in C, and block C is below block A.Then, a medium white triangle is below a A small red rectangle. A medium white triangle is below a A small red rectangle. Therefore, a small red rectangle is above a medium white triangle.A medium white rectangle is in A,a medium white triangle in C, and block C is below block A.Then, a medium white triangle is below a A medium white rectangle. A medium white triangle is below a A medium white rectangle. Therefore, a medium white rectangle is above a medium white triangle.A medium white rectangle is above a medium white triangle.Thus, all rectangles from given story are above a white triangle. Answer: Yes'

$\mathbf{user:}$ There is a block called A. Inside A, a black triangle above and to the left of a black circle. There is another block called B which is to the right of A. B has two yellow circles, a blue circle, and a blue square. The blue circle is touching the left edge of B. One of the yellow circles is to the right of the blue circle and to the left of the blue square. The blue square is touching the right edge of B and is above the other yellow circle, Which is also touching the right edge of B. Does block B have all of the circles inside it?

$\mathbf{assistant:}$ From given relation, A black triangle is above a black circle in A.Then, there is black circle in A.Therefore, not all circle are in B. Answer: No

$\mathbf{user:}$ There is a block called A. A has a large blue square near and below a medium blue triangle. Far to the left of the medium blue triangle is a small yellow square. Far above the medium blue triangle is a small blue triangle. There is another block called B, which is to the right of A. In B, There is a small yellow circle touching the left edge of B. Far to the right of the circle is a large yellow square. Near and above the large square is a small yellow square. There is another block called C, which is to the right of B. In C, there is a small blue circle near and to the left of a large yellow triangle. The large yellow triangle is near and to the left of a large blue triangle. The large blue triangle is far above a large blue circle. The large blue circle is touching the bottom edge of C. Are the large blue things in C to the left of the large blue thing in A?

$\mathbf{assistant:}$ Based on the provided content. Block B is to the right if block A.Block C is to the right of block B.Therefore, block C is to the right of block A.In block C, the yellow triangle is near and to the left of large blue triangle. Therefore, there is large blue triangle in C.Block A has a large blue square near and below a medium blue triangle.Therefore, there is large blue square in A.Then, large blue triangle in C is to the right of large blue square in A.So, large blue triangle in C is not to the left of large blue square in A. Answer: No

$\mathbf{user:}$ There is a block called A. In A there is a red triangle Which is above a black circle. The black circle is touching the bottom of A. A yellow square is touching the top edge of A, and below that, a red oval is touching the right edge of A. There is a second block call B Which is to the left of A. Inside B there are 3 squares, a brown one, a green one, and a red one. The green square is above the red square and touching the left edge of B. The brown square is touching both the bottom edge and the right edge of B. The red square is touching the left edge of the brown square. Is the oval to the right of the brown thing?

\subsection{Logical Representation (LR)}
$\mathbf{system:}$ You will be given story and question. Then, you will answer either Yes or No with explanation based on given story step by step. Candidate answer: [Yes, No]

$\mathbf{user:}$ There exist a big red square, a big red triangle, a medium red square, and a small green circle in a block called A. The triangle and medium square are touching the bottom edge of the block. The big and medium square are touching the right edge of the block. And the circle is above the big square which is above the medium square. There is another block called B to the left of block A. A medium green square is touching the left edge of block B and is below a medium red square. The medium red square is above and to the left of a small green square. Also a medium red triangle is below and to the left of the small square. Is the medium red triangle below the red square in block B?

$\mathbf{assistant:}$ Below(Medium red triangle in B, red square in block B) Answer: Yes

$\mathbf{user:}$ There are three blocks A, B and C. Block A has a medium white triangle and a small red rectangle. There exists a medium white rectangle to the left of the triangle and to the right of the red rectangle. Above block A there is block B. It has a small white oval touching the bottom edge of the block and a medium white triangle to the right of it. Below block A there is block C. Its has a small green triangle touching the top edge of the block. A medium white triangle is below a small green rectangle. Another small green triangle is below and to the right of the white triangle. Are all rectangles above a white triangle?'

$\mathbf{assistant:}$ Below(C, A) =$>$ Above(A, C), Contain(A, Small red rectangle) + Above(A, C) + Contain(C, Medium white triangle)=$>$Above(Small red rectangle, Medium White triangle), Above(A, C) \& Contain(A, Medium white rectangle) \& Contain(C, Medium white triangle)=$>$Above(Medium white rectangle, Medium White triangle), Below(Medium white triangle in C, small green rectangle in C) =$>$Above(small green rectangle, Medium white triangle)Answer: Yes

$\mathbf{user:}$ There is a block called A. Inside A, there is a black triangle that is above and to the left of a black circle. There is another block called B which is to the right of A. In B, there are two yellow circles, a blue circle, and a blue square. The blue circle is touching the left edge of B. One of the yellow circles is to the right of the blue circle and to the left of the blue square. The blue square is touching the right edge of B and is above the other yellow circle, Which is also touching the right edge of B. Does block B have all of the circles inside it?

$\mathbf{assistant:}$ Contain(A, Black Circle) =$>$ $\neg$Contain(B, Black Circle), $\neg$Contain(B, Black Circle) =$>$ Some x, $\neg$Contain(B, x) where x is Circle. Answer: No

$\mathbf{user:}$ There is a block called A. In A, there is a large blue square near and below a medium blue triangle. Far to the left of the medium blue triangle is a small yellow square. Far above the medium blue triangle is a small blue triangle. There is another block called B, which is to the right of A. In B, There is a small yellow circle touching the left edge of B. Far to the right of the circle is a large yellow square. Near and above the large square is a small yellow square. There is another block called C, which is to the right of B. In C, there is a small blue circle near and to the left of a large yellow triangle. The large yellow triangle is near and to the left of a large blue triangle. The large blue triangle is far above a large blue circle. The large blue circle is touching the bottom edge of C. Are the large blue things in C to the left of the large blue thing in A?

$\mathbf{assistant:}$ 
Question: Left(large blue thing in C, large blue thing in A). 
Related relation from story: Right(B, A), Right(C, B), Contain(C, large blue triangle), Contain(A, large blue square) 
Steps: Right(C, B) + Right(B, A) =$>$ Right(C, A) Contain(C, large blue triangle) + Right(C, A) + Contain(A, large blue square) =$>$ Right(large blue triangle in C, large blue square in A)Right(large blue triangle in C, large blue square in A) =$>$ !Left(large blue thing in C, large blue thing in A) Answer: No


\subsection{Chain of Symbol}
$\mathbf{system:}$  You will be given story and question. Then, you will answer either Yes or No with explanation based on given story step by step. Candidate answer: [Yes, No]

$\mathbf{user:}$ There exist a big red square, a big red triangle, a medium red square, and a small green circle in a block called A. The triangle and medium square are touching the bottom edge of the block. The big and medium square are touching the right edge of the block. And the circle is above the big square which is above the medium square. There is another block called B to the left of block A. A medium green square is touching the left edge of block B and is below a medium red square. The medium red square is above and to the left of a small green square. Also a medium red triangle is below and to the left of the small square. Is the medium red triangle below the red square in block B?

$\mathbf{assistant:}$ Let analyze the relations given from context to answer question.B: 1. (medium, red, triangle) $\downarrow$, $<$ (small, red, square)Based on the information, B: 1. (medium, red, triangle) $\downarrow$, $<$ (small, red, square)2. (medium, red, triangle) $\downarrow$ (small, red, square)Answer: Yes

$\mathbf{user:}$ There are three blocks A, B and C. Block A has a medium white triangle and a small red rectangle. There exists a medium white rectangle to the left of the triangle and to the right of the red rectangle. Above block A there is block B. It has a small white oval touching the bottom edge of the block and a medium white triangle to the right of it. Below block A there is block C. Its has a small green triangle touching the top edge of the block. A medium white triangle is below a small green rectangle. Another small green triangle is below and to the right of the white triangle. Are all rectangles above a white triangle?'

$\mathbf{assistant:}$ Let consider the related relations to answer the questions, A: 1. (small, red, rectangle) 2. (medium, white, rectangle) C: 1. (small, green, rectangle) = edge 2. (medium, white, triangle) ) $\downarrow$ (small, green, rectangle) C $\downarrow$ A Based on the information, let determined the relation of all rectangle and white triangles, 1. C $\downarrow$ A. Therefore, A $\uparrow$ C 2. A $\uparrow$ C, A: (small, red, rectangle), and C: (medium, white, triangle). Therefore, A: (small, red, rectangle) $\uparrow$ C: (medium, white, triangle) 3. A $\uparrow$ C, A: (medium, white, rectangle), and C: (medium, white, triangle). Therefore, A: (medium, white, rectangle) $\uparrow$ C: (medium, white, triangle). 4. C: (medium, white, triangle) ) $\downarrow$ (small, green, rectangle). Therefore, (small, green, rectangle) $\uparrow$ (medium, white, triangle).5. A: (small, red, rectangle) $\uparrow$ C: (medium, white, triangle), A: (medium, white, rectangle) $\uparrow$ C: (medium, white, triangle), and C: (small, green, rectangle) $\uparrow$ (medium, white, triangle), therefore, all rectangle above a white triangle. Answer: Yes

$\mathbf{user:}$ There is a block called A. Inside A, there is a black triangle that is above and to the left of a black circle. There is another block called B which is to the right of A. In B, there are two yellow circles, a blue circle, and a blue square. The blue circle is touching the left edge of B. One of the yellow circles is to the right of the blue circle and to the left of the blue square. The blue square is touching the right edge of B and is above the other yellow circle, Which is also touching the right edge of B. Does block B have all of the circles inside it?

$\mathbf{assistant:}$ Let consider the related relations to answer the questions,A: 1. (black, triangle) $\uparrow$ $<$ (black, circle)B: 1. (yellow, circle1) right (blue, circle)2. (blue, square) $\uparrow$ (yellow, circle2) 3.(blue, circle) =left edge of B Based on the information, let determined the relation of all rectangle and white triangles, 1. A: 1. (black, triangle) $\uparrow$$<$ (black, circle). Therefore, A: (black, circle)2. A: (black, circle). Therefore, B doesn't have all circle. Answer: No

$\mathbf{user:}$ There is a block called A. In A, there is a large blue square near and below a medium blue triangle. Far to the left of the medium blue triangle is a small yellow square. Far above the medium blue triangle is a small blue triangle. There is another block called B, which is to the right of A. In B, There is a small yellow circle touching the left edge of B. Far to the right of the circle is a large yellow square. Near and above the large square is a small yellow square. There is another block called C, which is to the right of B. In C, there is a small blue circle near and to the left of a large yellow triangle. The large yellow triangle is near and to the left of a large blue triangle. The large blue triangle is far above a large blue circle. The large blue circle is touching the bottom edge of C. Are the large blue things in C to the left of the large blue thing in A?

$\mathbf{assistant:}$ Let consider the related relations to answer the questions,B $>$ A,C $>$ BA: 1. (large, blue square) $\sim$ $\downarrow$ (medium, blue square)C: 1. (large, yellow, triangle)Based on the information, let determined the relation large blue things in C and the large blue thing in A,1. C $>$ B and B $>$ A. Therefore, C $>$ A2. A: 1. (large, blue, square) $\sim$$\downarrow$ (medium, blue square). Therefore, A: (large, blue, square)3. C $>$ A, A: (large, blue, square), and C: (large, yellow, triangle).Therefore, C: (large, yellow, triangle) $>$ A: (large, blue, square).4. C: (large, yellow, triangle) $>$ A: (large, blue, square).Therefore, C: (large, yellow, triangle) !$<$ A: (large, blue, square)large blue things in C is not to the left of the large blue thing in A. Answer: No

$\mathbf{user:}$ There is a block called A. In A there is a red triangle Which is above a black circle. The black circle is touching the bottom of A. A yellow square is touching the top edge of A and below that there is a red oval is touching the right edge of A. There is a second block call B Which is to the left of A. Inside B there are 3 squares, a brown one, a green one, and a red one. The green square is above the red square and touching the left edge of B. The brown square is touching both the bottom edge and the right edge of B. The red square is touching the left edge of the brown square. Is the oval to the right of the brown thing?

\subsection{Step-by-Step Prompting}

$\mathbf{Raw \, context:}$ There exist a big red square, a big red triangle, a medium red square, and a small green circle in a block called A. The triangle and medium square are touching the bottom edge of the block. The big and medium square are touching the right edge of the block. And the circle is above the big square which is above the medium square. There is another block called B to the left of block A. A medium green square is touching the left edge of block B and is below a medium red square. The medium red square is above and to the left of a small green square. Also a medium red triangle is below and to the left of the small square.

$\mathbf{Spliting \, context:}$  A big red square in block A.

A big red triangle in block A.

A medium red square in block A.

A small green circle in block A.

The triangle is touching the bottom edge of block A.

The medium square is touching the bottom edge of block A.

The big square is touching the right edge of block A.

The medium square is touching the right edge of block A.

The circle is above the big square in A.

The big square is above the medium square in A.

Block B is to the left of block A.

A medium green square is touching the left edge of block B.

A medium green square is below a medium red square in B.

The medium red square is above a small green square in B.

The medium red square is left of a small green square in B.

The medium red triangle is below the small square in B.

The medium red triangle is left the small square in B.

\section{Spatial Logical Rules}
\label{appendix:spatial_logic_rules}
The conversion from spatial logical rules proposed in ~\cite{mirzaee2022transfer} to logical constraints used in our experiment is shown in Table~\ref{tab:rule_table_ex}.

\begin{table*}[ht!]
    \begin{adjustbox}{width=\columnwidth*2 + 5mm, center}
    \begin{tabular}{ |c | c | c | } 
     \hline
     Rule Type & Rule & Logical Constraints\\
     \hline
& $above(x, y) \Rightarrow below(y, x)$ &   \\ 
& $below(x, y) \Rightarrow above(y, x)$ &   \\
& $left(x, y) \Rightarrow right(y, x)$ &   \\ 
& $right(x, y) \Rightarrow left(y, x)$ &   \\
& $front(x, y) \Rightarrow behind(y, x)$ &   \\ 
Converse & $behind(x, y) \Rightarrow front(y, x)$ &  $Truth(q_i) \Rightarrow Truth(q_j) $  \\
& $coveredby(x, y) \Rightarrow cover(y, x)$ &   \\ 
& $cover(x, y) \Rightarrow coveredby(y, x)$ &   \\
& $inside(x, y) \Rightarrow contain(y, x)$ &   \\ 
& $contain(x, y) \Rightarrow inside(y, x)$ &   \\
\hline
& $near(x, y) \Rightarrow near(y, x)$ &   \\ 
& $far(x, y) \Rightarrow far(y, x)$ &   \\ 
Symmetric & $touch(x, y) \Rightarrow touch(y, x)$ & $Truth(q_i) \Rightarrow Truth(q_j)$  \\ 
& $disconnected(x, y) \Rightarrow disconnected(y, x)$ &   \\ 
& $overlap(x, y) \Rightarrow overlap(y, x)$ &   \\ 
\hline
& $left(x, y) \land left(y, z) \Rightarrow left(x, z)$ &    \\
& $right(x, y) \land right(y, z) \Rightarrow right(x, z)$ &    \\
& $above(x, y) \land above(y, z) \Rightarrow above(x, z)$ &    \\
& $below(x, y) \land below(y, z) \Rightarrow below(x, z)$ &    \\
& $behind(x, y) \land behind(y, z) \Rightarrow behind(x, z)$ &    \\
& $front(x, y) \land front(y, z) \Rightarrow front(x, z)$ &    \\
& $inside(x, y) \land inside(y, z) \Rightarrow inside(x, z)$ &    \\
& $contain(x, y) \land contain(y, z) \Rightarrow contain(x, z)$ &    \\
& $inside(x, y) \land coveredby(y, z) \Rightarrow inside(x, z)$ &    \\
& $contain(x, y) \land cover(y, z) \Rightarrow contain(x, z)$ &    \\
& $inside(x, y) \land left(y, z) \Rightarrow left(x, z)$ &    \\
& $inside(x, y) \land right(y, z) \Rightarrow right(x, z)$ &    \\
& $inside(x, y) \land above(y, z) \Rightarrow above(x, z)$ &    \\
& $inside(x, y) \land below(y, z) \Rightarrow below(x, z)$ &    \\
Transitivity & $inside(x, y) \land behind(y, z) \Rightarrow behind(x, z)$ & $Truth(q_i) \land Truth(q_j) \Rightarrow Truth(q_k)$   \\
& $inside(x, y) \land front(y, z) \Rightarrow front(x, z)$ &    \\
& $inside(x, y) \land near(y, z) \Rightarrow near(x, z)$ &    \\
& $inside(x, y) \land far(y, z) \Rightarrow far(x, z)$ &    \\
& $inside(x, y) \land disconnected(y, z) \Rightarrow disconnected(x, z)$ &    \\
& $coveredby(x, y) \land left(y, z) \Rightarrow left(x, z)$ &    \\
& $coveredby(x, y) \land right(y, z) \Rightarrow right(x, z)$ &    \\
& $coveredby(x, y) \land above(y, z) \Rightarrow above(x, z)$ &    \\
& $coveredby(x, y) \land below(y, z) \Rightarrow below(x, z)$ &    \\
& $coveredby(x, y) \land behind(y, z) \Rightarrow behind(x, z)$ &    \\
& $coveredby(x, y) \land front(y, z) \Rightarrow front(x, z)$ &    \\
& $coveredby(x, y) \land near(y, z) \Rightarrow near(x, z)$ &    \\
& $coveredby(x, y) \land far(y, z) \Rightarrow far(x, z)$ &    \\
& $coveredby(x, y) \land disconnected(y, z) \Rightarrow disconnected(x, z)$ &    \\
\hline
     \end{tabular}
    \end{adjustbox}
    \caption{The conversion from spatial logical rules proposed in ~\cite{mirzaee2022transfer} to logical constraints used in our experiment.}
    \label{tab:rule_table_ex}
\end{table*}

\newpage

\section{DomiKnowS Concepts and Graphs}\label{appendix:domiknows}
Graph and concepts declaration of our work inside the DomiKnowS framework.

\subsection{Graph for YN}

\begin{lstlisting}[language=Python]
story = Concept(name="story")
question = Concept(name="question")
story_contain, = story.contains(question)
answer_class = question(name="answer_class", ConceptClass=EnumConcept,
                        values=["yes", "no"])
symmetric = Concept(name="symmetric")
s_quest1, s_quest2 = symmetric.has_a(arg1=question, arg2=question)
ifL(andL(answer_class.yes('x'), existsL(symmetric('s', path=('x', symmetric)))),
    answer_class.yes(path=('s', s_quest2)))

reverse = Concept(name="reverse")
r_quest1, r_quest2 = reverse.has_a(arg10=question, arg20=question)
ifL(andL(answer_class.yes('x'), existsL(reverse('r', path=('x', reverse)))),
    answer_class.no(path=('r', r_quest2)))
ifL(andL(answer_class.no('x'), existsL(reverse('r', path=('x', reverse)))),
    answer_class.yes(path=('r', r_quest2)))

transitive = Concept(name="transitive")
t_quest1, t_quest2, t_quest3 = transitive.has_a(arg11=question, arg22=question, arg33=question)
ifL(andL(answer_class.yes('x'), existsL(transitive("t", path=('x', transitive))), answer_class.yes(path=('t', t_quest2))),
    answer_class.yes(path=('t', t_quest3)))

transitive_topo = Concept(name="transitive_topo")
tt_quest1, tt_quest2, tt_quest3, tt_quest4 = transitive_topo.has_a(arg111=question, arg222=question,arg333=question, arg444=question)
ifL(andL(answer_class.yes('x'), existsL(transitive("t", path=('x', transitive_topo))),
         answer_class.yes(path=('t', tt_quest2)), answer_class.yes(path=('t', tt_quest3))),
    answer_class.yes(path=('t', tt_quest4)))
\end{lstlisting}

\subsection{Graph for FR on SpaRTUN}
\begin{lstlisting}[language=Python]
story = Concept(name="story")
question = Concept(name="question")
story_contain, = story.contains(question)

left = question(name="left")
right = question(name="right")
above = question(name="above")
below = question(name="below")
behind = question(name="behind")
front = question(name="front")
near = question(name="near")
far = question(name="far")
disconnected = question(name="disconnected")
touch = question(name="touch")
overlap = question(name="overlap")
coveredby = question(name="coveredby")
inside = question(name="inside")
cover = question(name="cover")
contain = question(name="contain")
output_for_loss = question(name="output_for_loss")

# Only one label of opposite concepts
exactL(left, right)
exactL(above, below)
exactL(behind, front)
exactL(near, far)
exactL(disconnected, touch)

# Inverse Constrains
inverse = Concept(name="inverse")
inv_question1, inv_question2 = inverse.has_a(arg1=question, arg2=question)

# First inverse relation, allow inverse back and forth
inverse_list1 = [(above, below), (left, right), (front, behind), (coveredby, cover),
                (inside, contain)]

for ans1, ans2 in inverse_list1:
    ifL(andL(ans1('x'), existsL(inverse('s', path=('x', inverse)))),
        ans2(path=('s', inv_question2)))

    ifL(andL(ans2('x'), existsL(inverse('s', path=('x', inverse)))),
        ans1(path=('s', inv_question2)))

# 2 PMD : = entropy + beta * constraint_loss ( Train with no-constraint first then working on)
# symmetric
inverse_list2 = [(near, near), (far, far), (touch, touch), (disconnected, disconnected), (overlap, overlap)]
for ans1, ans2 in inverse_list2:
    ifL(andL(ans1('x'), existsL(inverse('s', path=('x', inverse)))),
        ans2(path=('s', inv_question2)))

# Transitive constrains
transitive = Concept(name="transitive")
tran_quest1, tran_quest2, tran_quest3 = transitive.has_a(arg11=question, arg22=question, arg33=question)

transitive_1 = [left, right, above, below, behind, front, inside, contain]

for rel in transitive_1:
    ifL(andL(rel('x'),
             existsL(transitive("t", path=('x', transitive))),
             rel(path=('t', tran_quest2))),
        rel(path=('t', tran_quest3)))
# Transitive of cover and contain
transitive_2 = [(coveredby, inside), (cover, contain)]
for rel1, rel2 in transitive_2:
    ifL(andL(rel2('x'),
             existsL(transitive("t", path=('x', transitive))),
             rel1(path=('t', tran_quest2))),
        rel2(path=('t', tran_quest3)))

# Transitive of inside/cover with position
transitive_3_1 = [inside, coveredby]
transitive_3_2 = [left, right, above, below, behind, front, near, far, disconnected]
for rel1 in transitive_3_1:
    for rel2 in transitive_3_2:
        ifL(andL(rel1('x'),
                 existsL(transitive("t", path=('x', transitive))),
                 rel2(path=('t', tran_quest2))),
            rel2(path=('t', tran_quest3)))

# Transitive + topo constrains
tran_topo = Concept(name="transitive_topo")
tran_topo_quest1, tran_topo_quest2, tran_topo_quest3, tran_topo_quest4 = tran_topo.has_a(arg111=question, arg222=question, arg333=question, arg444=question)
# (x inside y) + (h inside z) + (y direction z) => (x direction h)
tran_topo_2_1 = [inside, coveredby]
tran_topo_2_2 = [left, right, above, below, behind, front, near, far, disconnected]
for rel1 in tran_topo_2_1:
    for rel2 in tran_topo_2_2:
        ifL(andL(rel1('x'),
                 existsL(tran_topo('to', path=('x', tran_topo))),
                 rel1(path=('to', tran_topo_quest2)),
                 rel2(path=('to', tran_topo_quest3))
                 ),
            rel2(path=('to', tran_topo_quest4)))

tran_topo_3_1 = [left, right, above, below, behind, front, near, far, disconnected]
tran_topo_3_2 = [contain, cover]
for rel1 in tran_topo_3_1:
    for rel2 in tran_topo_3_2:
        ifL(andL(rel1('x'),
                 existsL(tran_topo('to', path=('x', tran_topo))),
                 rel1(path=('to', tran_topo_quest2)),
                 rel2(path=('to', tran_topo_quest3))),
            rel1(path=('to', tran_topo_quest4)))
\end{lstlisting}

\subsection{Graph for FR on StepGame}
\begin{lstlisting}[language=Python]
story = Concept(name="story")
question = Concept(name="question")
story_contain, = story.contains(question)

left = question(name="left")
right = question(name="right")
above = question(name="above")
below = question(name="below")
lower_left = question(name="lower-left")
lower_right = question(name="lower_right")
upper_left = question(name="upper_left")
upper_right = question(name="upper_right")
overlap = question(name="overlap")
\end{lstlisting}

\end{document}